\documentclass[review]{elsarticle}
\usepackage{lineno}
\usepackage[hidelinks]{hyperref}

\usepackage{multirow,setspace,amssymb,amsmath,graphicx,color,rotating,subfigure,url}
\usepackage{natbib}
\setcitestyle{authoryear,longnamesfirst,open={(},close={)}} 
\setcitestyle{citepsep={;}}
\usepackage{textcomp}
\usepackage{CJK}
\usepackage{bm}%
\usepackage{rotating}
\usepackage{epsfig}
\usepackage{dcolumn}
\usepackage{epstopdf}

\usepackage{graphicx}
\usepackage{dcolumn}
\usepackage{bm}
\usepackage{color}
\usepackage{amsmath}
\usepackage[utf8]{inputenc}
\usepackage{doi}
\usepackage[T1]{fontenc}
\usepackage{diagbox}
\usepackage{amssymb}
\usepackage{enumerate}
\usepackage[ruled,linesnumbered]{algorithm2e}
\usepackage{amsmath}
\usepackage{mathrsfs}
\usepackage{booktabs}
\usepackage[ruled]{algorithm2e}

\usepackage{indentfirst}
\usepackage{xcolor}

\begin{document}
\begin{frontmatter}

\title{Directing the Narrative: A Finetuning Method for Controlling Coherence and Style in Story Generation  }

\author[hznu]{Jianzhang Zhang}
\ead{zjzhang@hznu.edu.cn}

\author[hznu]{Yijing Tian}
\ead{xxx@hznu.edu.cn}

\author[hznu]{Jiwang Qu}
\ead{xxx@hznu.edu.cn}


\author[hznu]{Chuang Liu\corref{cor1}}
\ead{liuchuang@hznu.edu.cn}

\cortext[cor1]{Corresponding author.}

\affiliation[hznu]{
	organization={Department of Management Science and Engineering, Hangzhou Normal University},
	city={Hangzhou},
	state={Zhejiang},
	country={P.R. China}
}



\begin{abstract}
Story visualization requires generating sequential imagery that aligns semantically with evolving narratives while maintaining rigorous consistency in character identity and visual style. However, existing methodologies often struggle with subject inconsistency and identity drift, particularly when depicting complex interactions or extended narrative arcs. To address these challenges, we propose a cohesive two-stage framework designed for robust and consistent story generation. First, we introduce Group-Shared Attention (GSA), a mechanism that fosters intrinsic consistency by enabling lossless cross-sample information flow within attention layers. This allows the model to structurally encode identity correspondence across frames without relying on external encoders. Second, we leverage Direct Preference Optimization (DPO) to align generated outputs with human aesthetic and narrative standards. Unlike conventional methods that rely on conflicting auxiliary losses, our approach simultaneously enhances visual fidelity and identity preservation by learning from holistic preference data. Extensive evaluations on the ViStoryBench benchmark demonstrate that our method establishes a new state-of-the-art, significantly outperforming strong baselines with gains of +10.0 in Character Identity (CIDS) and +18.7 in Style Consistency (CSD), all while preserving high-fidelity generation.
\end{abstract}


\end{frontmatter}

\section{Introduction}
\label{sec:introduction}

\begin{figure}[t]
    \centering
    \includegraphics[width=1\linewidth]{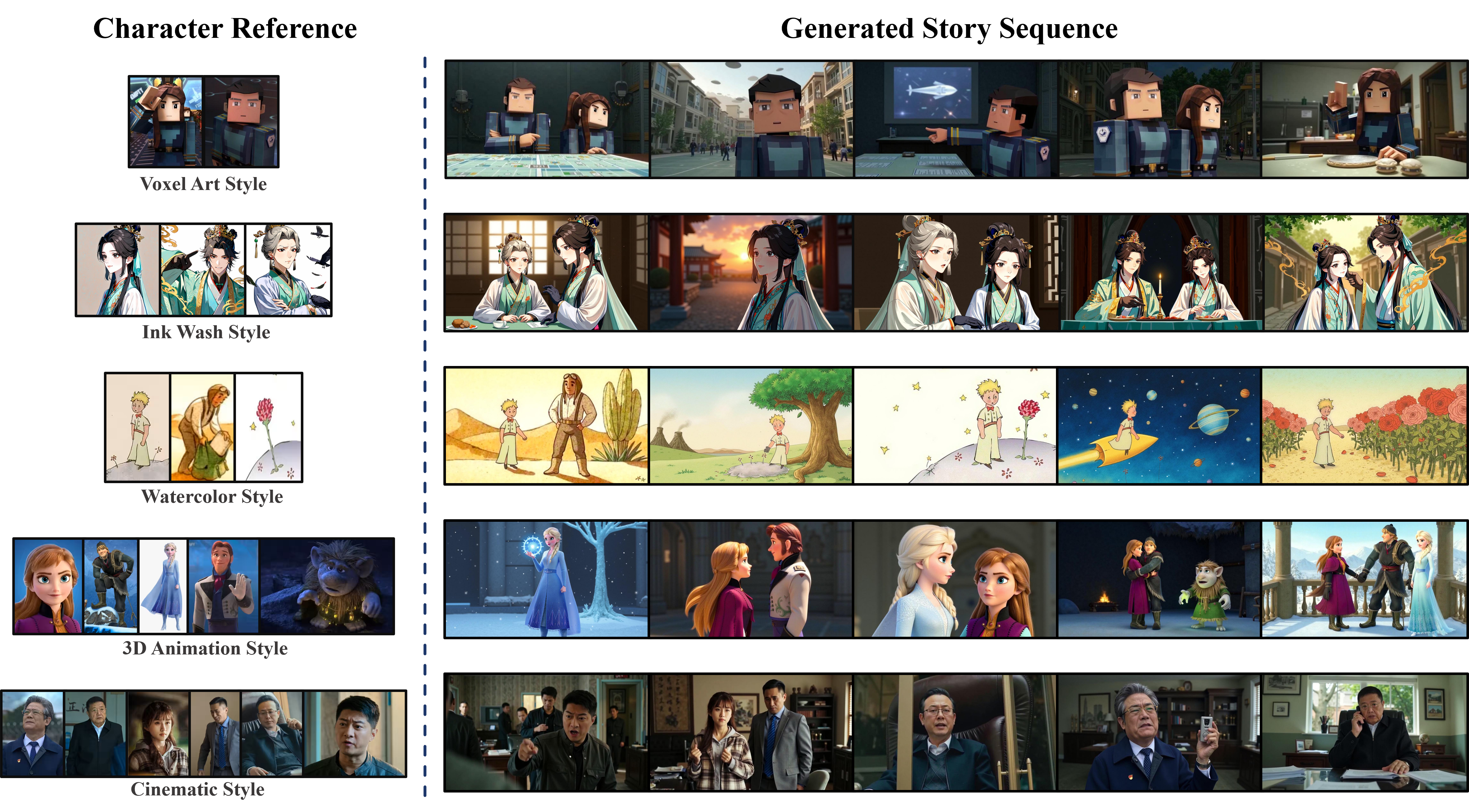}
    \caption{Visual showcase of consistent character generation. The leftmost column displays the reference input. Subsequent columns demonstrate our model's capability to synthesize the same character in various complex scenarios, maintaining strictly consistent identity and style.}
    \label{showcase}
\end{figure}

Transforming textual narratives into coherent visual sequences stands as a pivotal challenge in computer vision and text-to-image synthesis. Story visualization demands that a model function as a consistent visual narrator, synthesizing sequential frames that faithfully align with evolving plot descriptions while rigorously preserving character identity and artistic style. This task presents an inherent trade-off between generative diversity and identity consistency: the model must possess the flexibility to render varied scenes and poses without compromising the structural stability required to prevent the protagonist’s appearance from drifting.

While recent diffusion transformers (DiTs) demonstrate exceptional visual fidelity, applying them directly to story visualization remains challenging. Since standard diffusion models are trained on independent samples, they lack an intrinsic mechanism to maintain correspondence across a sequence. Consequently, when generating a multi-frame story, these models frequently suffer from identity drift, in which a character's facial features or clothing gradually mutate. Furthermore, they often exhibit style inconsistency that disrupts the narrative flow~\citep{zhou2024storydiffusion}.

To mitigate these issues, existing methodologies typically rely on two primary paradigms. The first, optimization-based methods such as LoRA~\citep{hu2022lora} and DreamBooth~\citep{ruiz2023dreambooth} fine-tune weights for specific characters. While effective, these approaches necessitate computationally intensive per-subject optimization, severely limiting their scalability. The second paradigm, encoder-based methods exemplified by IP-Adapter~\citep{ye2023ipadapter}, utilizes external vision encoders (e.g., CLIP) to condense reference images into feature embeddings, which are then injected via cross-attention. However, this reliance on external compression often creates an information bottleneck, resulting in the loss of fine-grained identity details and inducing semantic conflict with textual controls.

We argue that consistency should be an intrinsic capability learned through direct feature interaction, rather than relying on compressed external injections. To this end, we propose a novel framework that transforms the pre-trained FLUX.1\footnote{\url{https://github.com/black-forest-labs/flux}} backbone into a consistent story visualizer through a progressive two-stage training strategy.

In the first stage, we introduce Group-Shared Attention (GSA). Departing from encoder-based paradigms that depend on compressed global embeddings, GSA facilitates lossless internal information flow. By repurposing the self-attention mechanism to share high-resolution visual tokens across images within the same training batch, GSA enables the model to directly attend to the spatial features of reference views. This design ensures fine-grained structural consistency in identity and attire without compromising the model's adherence to textual prompts.

While GSA establishes structural correspondence, standard reconstruction objectives may still yield anatomical artifacts. Therefore, in the second stage, we employ Direct Preference Optimization (DPO) to align the model with human aesthetic standards. Using an expert-curated dataset, we train the model to discriminate between consistent, high-fidelity generations and those exhibiting subtle distortions. This stage functions as a semantic refinement process, enhancing perceptual quality and reinforcing identity constraints. Figure \ref{showcase} demonstrates the final coherent narratives achieved by our framework.

Our contributions are summarized as follows:
\begin{itemize}
    \item We propose Group-Shared Attention (GSA), a novel mechanism that enables structured, lossless information flow across samples. Distinct from encoder-based injection, GSA empowers the model to intrinsically learn robust identity preservation directly from high-resolution internal features.
    
    \item We introduce a holistic two-stage training framework that harmonizes structural consistency with human aesthetic preferences. By combining structural pre-training with DPO-based refinement, our strategy effectively balances the trade-off between rigid identity adherence and high-fidelity semantic generation.
    
    \item We achieve new state-of-the-art performance on the ViStoryBench benchmark~\citep{zhuang2025vistorybench}. Extensive experiments demonstrate significant improvements of +10.0 in Character Identity (CIDS) and +18.7 in Style Consistency (CSD) compared to strong baselines, setting a new standard for consistent story visualization.
\end{itemize}

\section{Related Works} \label{Related works}

\subsection{Text-to-Image Generation}
Text-to-image synthesis has evolved from early Generative Adversarial Networks~\citep{goodfellow2020generative} to diffusion-based frameworks.
While GANs provided initial solutions, they often faced challenges regarding training stability.
Denoising Diffusion Probabilistic Models subsequently offered improved stability and generation fidelity.
Modern models have further scaled up image synthesis capabilities.
DALL-E~2~\citep{ramesh2022hierarchical} and Imagen~\citep{saharia2022photorealistic} demonstrated the effectiveness of leveraging large text encoders.
Latent Diffusion Models~\citep{rombach2022high}, such as Stable Diffusion, introduced the efficiency of latent space processing, while subsequent iterations like SDXL~\citep{podell2023sdxl} and Pixart-\(\alpha\)~\citep{chen2023pixart} increased model capacity.

Most recently, the field has progressed towards Flow Matching combined with Diffusion Transformer architectures.
Models such as FLUX.1 leverage rectified flow matching to learn straight transport paths between noise and data.
These flow-based transformers offer improved prompt adherence compared to earlier architectures and serve as the foundational backbone for our investigation.

\subsection{Image Customization}
Image customization aims to synthesize specific subjects based on reference images.
Optimization-based methods like Textual Inversion~\citep{gal2022image} and DreamBooth~\citep{ruiz2023dreambooth} fine-tune embeddings or model parameters.
To reduce computational costs, Parameter-Efficient Fine-Tuning methods like LoRA~\citep{hu2022lora} were developed to optimize a subset of parameters.
Zero-shot methods such as IP-Adapter~\citep{ye2023ipadapter} and InstantID~\citep{wang2024instantid} utilize encoder-based mechanisms to inject identity features directly.

Addressing the complexity of multi-subject generation, UNO~\citep{wu2025less} introduces a Model-Data Co-evolution strategy to construct a sophisticated pipeline for generating aligned data. While these approaches have achieved significant success, there remains a need to investigate methods that directly learn from open-domain natural images without relying on extensive synthetic data construction, offering a complementary perspective to existing pipelines.

\subsection{Story Visualization}
Story visualization requires generating sequences of frames that align with a narrative.
The primary task is to maintain contextual consistency across time steps.
Early RNN-based methods like StoryGAN~\citep{li2019storygan} and autoregressive diffusion approaches often faced stability issues.
StoryGen~\citep{liu2023intelligent} improved this by defining an open-ended task utilizing a Visual-Language Context Module, though autoregressive models still contend with error accumulation over longer sequences.

To improve consistency, recent works have utilized attention modification.
StoryDiffusion~\citep{zhou2024storydiffusion} applies a consistent self-attention mechanism to share features across frames, while Story-Adapter~\citep{mao2024story} employs a Global Reference Cross-Attention module for iterative refinement.
However, these mechanisms have been primarily tailored for U-Net-based architectures.
With the emergence of Rectified Flow Transformers, adapting such consistency strategies to this new paradigm presents a distinct challenge.
Our work addresses this by exploring the implementation of identity preservation mechanisms specifically within a flow-based architecture.

\section{Preliminaries and Data Construction} 
\label{sec:dataset}

In this section, we first formulate the problem of character-consistent story generation and then detail the systematic data curation pipeline designed to support this task.

\begin{figure}[htbp] 
    \centering
    \includegraphics[width=1\linewidth]{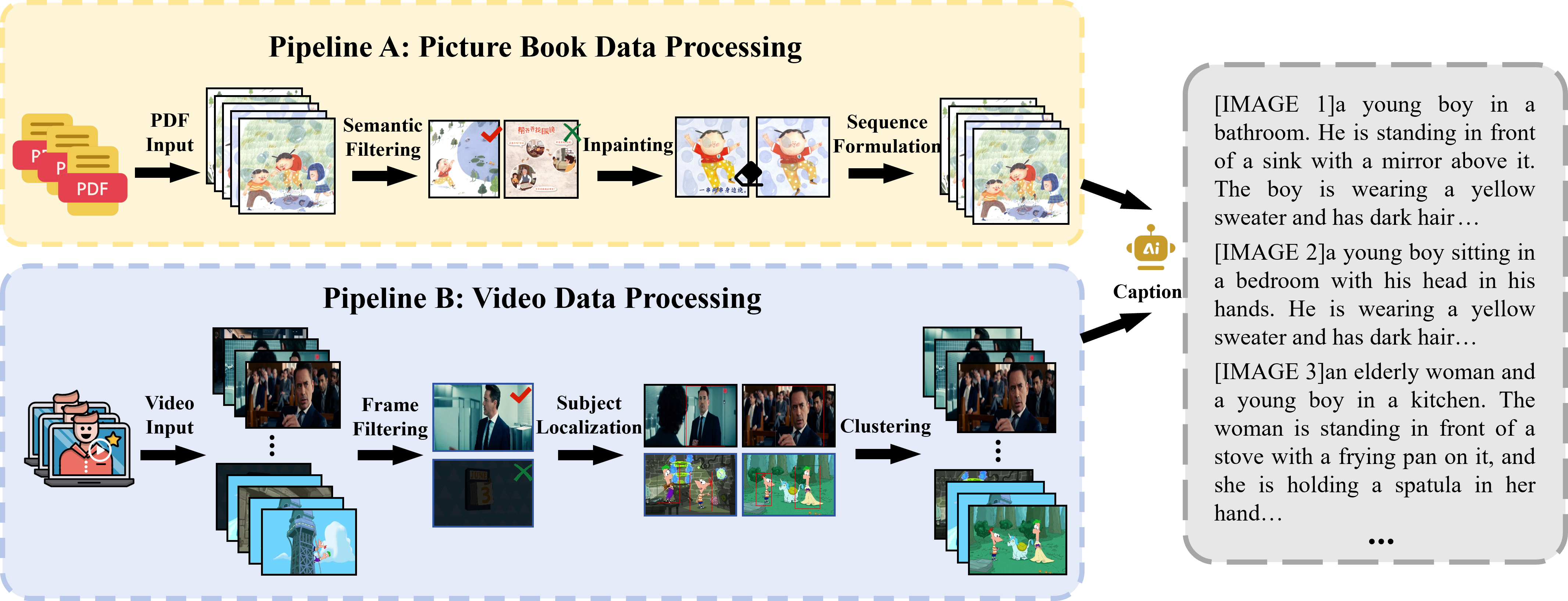}
    \caption{Illustration of the data construction process. The pipeline transforms raw data from storybooks and films into character-consistent sequences.}
    \label{data_construction}
\end{figure}
\subsection{Task Definition}
\label{subsec:task_def}
Formally, let $S = \{t_1, t_2, ..., t_N\}$ denote a sequence of textual narratives describing a continuous story. Our objective is to generate a corresponding image sequence $I = \{x_1, x_2, ..., x_N\}$. The generation process is constrained by two objectives: (1) \textit{Semantic Alignment}, where each image $x_i$ must accurately reflect the prompt $t_i$; and (2) \textit{Identity Consistency}, requiring the character identity to remain visually invariant across all frames $x_i \in I$.

\subsection{Pipeline A: Picture Book Data Processing}
We initially aggregated an extensive collection of over 10,000 open-source illustrated storybooks. To transform these raw PDF documents into a high-fidelity, training-ready dataset, we established a systematic three-stage preprocessing workflow.

\paragraph{Semantic Filtering}
Raw inputs often contain significant noise, including book covers, prefaces, and non-narrative layouts. We developed a custom semantic filter by fine-tuning a CLIP model on a manually labeled subset of 1,000 narrative versus non-narrative pages.We enforced a strict confidence threshold of 0.9, discarding all samples below this value to ensure narrative relevance. We further applied a resolution constraint requiring both height and width to exceed 512 pixels. Due to these rigorous quality standards, the candidate pool was refined to approximately 12,000 high-quality images.

\paragraph{Structure-Aware Inpainting}
Text embedded in illustrations hinders the model from learning pure visual representations. We employed an OCR-based detection module to localize text regions, followed by a structure-aware inpainting model to reconstruct the occluded background seamlessly.

\paragraph{Sequence Formulation}
To support story-level training, we grouped images by their source books. After removing near-duplicates via perceptual hashing, we discarded books with fewer than 4 coherent frames. This process yielded 4,000 clean and text-free narrative images tailored for consistency learning.

\subsection{Pipeline B: Video Data Processing}
While video sources offer rich character dynamics, they present significant challenges regarding identity tracking and scene complexity. To address this, we processed a diverse collection of over 100 video sequences through a sequential pipeline designed to extract high-fidelity character data.

\paragraph{Hierarchical Frame Filtering}
To guarantee optimal visual fidelity from the outset, we enforced a strict filtering protocol immediately following shot boundary detection. For animated content, perceptual hashing was applied to eliminate redundant static frames. For live-action footage, we implemented specific exclusion criteria to mitigate common visual artifacts; frames exhibiting harsh lighting, motion blur, extreme long shots, or profile views were explicitly discarded.

\paragraph{Subject Localization}
Following filtration, GroundingDINO was utilized to localize characters within the curated frames. This step effectively isolates candidate subjects from complex background scenes, establishing a precise spatial baseline for subsequent identity analysis.

\paragraph{Coarse-to-Fine Character Clustering}
We established identity consistency via a coarse-to-fine strategy. Initially, DINOv2 features coupled with Louvain clustering formed candidate character groups, followed by object detection for contextual retrieval. Finally, Qwen-VL served as a semantic verifier to rigorously filter outliers, ensuring strict identity consistency. This pipeline ultimately yielded 300 high-consistency character clusters comprising approximately 1,500 premium frames.

\subsection{Automated Semantic Annotation}
Accurate text-image alignment requires dense and descriptive captions to serve as training prompts. We employed the Florence-2-large vision-language model in its detailed captioning mode. Unlike standard captioners which often miss fine-grained attributes, Florence-2 excels in recognizing spatial relationships and intricate character details. This provides the granular textual guidance necessary for our first training stage.

\section{Methodology}\label{Methodology}
\subsection{Method Overview}
To address the sequence generation task defined in Sec.~\ref{subsec:task_def}, we decompose the problem into a reference-conditioned generation framework. Instead of generating the entire sequence jointly, we treat consistency as a retrieval task where the current generation is conditioned on a set of reference character images.

\begin{figure}[t] 
    \centering
    \includegraphics[width=1\linewidth]{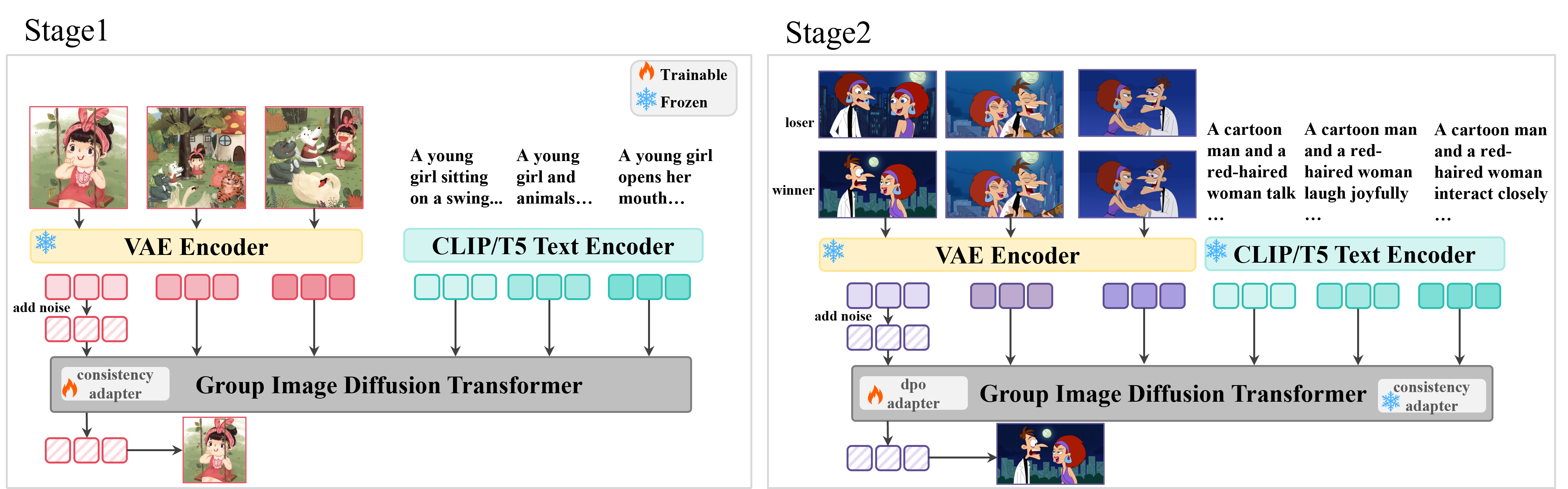}
    \caption{The pipeline of our two-stage approach.
    Stage 1 performs Consistency Pre-training by injecting Group-Shared Attention (GSA) into the adapter $\Phi^c$. 
    Stage 2 executes Preference Alignment via DPO (adapter $\Phi^d$) to mitigate semantic drift and enhance fidelity.}
    \label{fig:frame}
\end{figure}

As illustrated in Figure \ref{fig:frame}, we propose a progressive two-stage training strategy to inject identity consistency into the pre-trained FLUX.1 backbone.
\begin{itemize}
    \item Stage 1: We introduce a lightweight adapter $\Phi^c$ equipped with Group-Shared Attention (GSA). This module enables the model to implicitly learn identity consistency by retrieving high-resolution features from reference samples within the batch.
    \item Stage 2: To bridge the gap between structural consistency and human perceptual standards, we freeze the Stage 1 model and train an auxiliary adapter $\Phi^d$ via Direct Preference Optimization (DPO), aligning the output with high-fidelity human aesthetics.
\end{itemize}

The final model $M$ is a composition of the frozen backbone and the learned adapters:
\begin{equation}
    M = \Phi + \Phi^c + \Phi^d
\end{equation}

\subsection{Backbone and Flow Matching Formulation}
We utilize FLUX.1-dev as our backbone $\Phi$ due to its superior prompt adherence and high-fidelity VAE compression. Built upon the Rectified Flow framework~\citep{peebles2023scalable}, FLUX.1 learns to straighten the transport path between noise and data distributions. We adopt its standard flow matching objective, which serves as the foundation for our subsequent optimization:
\begin{equation}
\mathcal{L} = \mathbb{E}_{t, \mathbf{z}_t, \mathbf{c}} \left[ \| \mathbf{v}_t - \mathbf{v}_\theta(\mathbf{z}_t, t, \mathbf{c}) \|^2 \right]
\end{equation}
where $\mathbf{v}_\theta$ denotes the velocity field predicted by the network. We omit detailed architectural descriptions as they follow standard DiT implementations.

\subsection{Stage 1: Identity-Preserving Context Learning}

\subsubsection{Group-Shared Attention (GSA)}
\paragraph{Unified Internal Retrieval} Standard diffusion models generate images in isolation, synthesizing each sample solely from independent noise. This isolation precludes identity consistency across separate generation sessions. 
To address this, we formulate identity preservation as a Reference-Anchored retrieval task. We hypothesize that the generative Transformer can learn to maintain identity by directly attending to the high-resolution spatial features of reference images within the same batch. Instead of relying on compressed external encoders (e.g., CLIP), we leverage the internal attention mechanism of the model to retrieve identity details.

\paragraph{Cross-Sample Context Expansion} In standard DiT implementations, self-attention is restricted to the local scope of a single image. We reformulate this by expanding the context window. 
Given the asymmetric nature of our training strategy, we focus on the attention computation for the \textit{target} sample (indexed as $0$). The receptive field is expanded to incorporate visual tokens from all \textit{reference} samples (indexed as $j \in \{1, \dots, N-1\}$).

To prevent semantic conflict, we employ asymmetric multi-modal sharing: the target retains its full text-image context, while only visual tokens are retrieved from the references.
Formally, the extended key $\tilde{\mathbf{K}}^{(0)}$ and value $\tilde{\mathbf{V}}^{(0)}$ for the target sample are constructed as:
\begin{equation}
\tilde{\mathbf{K}}^{(0)} = [\mathbf{K}_{\text{txt}}^{(0)}, \mathbf{K}_{\text{img}}^{(0)}] \oplus \bigcup_{j=1}^{N-1} [\mathbf{K}_{\text{img}}^{(j)}]
\end{equation}
\begin{equation}
\tilde{\mathbf{V}}^{(0)} = [\mathbf{V}_{\text{txt}}^{(0)}, \mathbf{V}_{\text{img}}^{(0)}] \oplus \bigcup_{j=1}^{N-1} [\mathbf{V}_{\text{img}}^{(j)}]
\end{equation}
Subsequently, the attention for the target is computed as:
\begin{equation}
\text{GSA}(\mathbf{h}^{(0)}) = \text{softmax}\left(\frac{\mathbf{Q}^{(0)} (\tilde{\mathbf{K}}^{(0)})^{\top}}{\sqrt{d}}\right) \tilde{\mathbf{V}}^{(0)}
\end{equation}
Here, $\oplus$ denotes concatenation along the sequence dimension. Through this mechanism, the query $\mathbf{Q}^{(0)}$ learns to extract consistent identity features from the extended context $\tilde{\mathbf{K}}^{(0)}$.

\subsubsection{Training Strategy and Objective}

GSA is integrated into the FLUX.1 architecture via Low-Rank Adaptation (LoRA) applied to attention projections, feed-forward networks, and normalization layers. To effectively train this mechanism, we introduce a specialized sampling strategy.

\paragraph{Asymmetric Timestep Sampling} To align the training with the inference requirement of high-fidelity reference guidance, we introduce an asymmetric noise schedule within the batch. 
Specifically, for a training batch of size $N$, we designate the first sample (index 0) as the \textit{target} and the remaining $N-1$ samples as \textit{references}.
We assign a random diffusion timestep $t \sim \text{Sigmoid}(\mathcal{N}(0,1))$ to the target to induce noise, while strictly enforcing $t=0$ for all reference samples. 
This ensures that the key-value pairs provided by the references contain clean, noise-free visual features, effectively creating a stable visual context pool for the target to query during the denoising process.

\paragraph{Training Objective} Consequently, the flow matching loss is computed exclusively on the target sample. The model is optimized to predict the velocity vector of the target, conditioned on the clean reference latents:
\begin{equation}
\mathcal{L}_{\text{stage1}} = \mathbb{E}_{t, \mathbf{z}_0, \epsilon} \left[ \left\| \mathbf{v}_t^{(0)} - \mathbf{v}_\theta(\mathbf{z}_t^{(0)}, t, \mathbf{c}^{(0)}, \{\mathbf{z}_0^{(j)}\}_{j=1}^{N-1}) \right\|^2 \right]
\end{equation}
where $\mathbf{z}_t^{(0)}$ is the noisy target latent, and $\{\mathbf{z}_0^{(j)}\}$ represents the set of clean reference latents accessible via GSA. This objective forces the attention adapter to learn how to utilize clean reference details to reconstruct the target identity.

\subsection{Stage 2: Refinement via Direct Preference Optimization}

While Stage 1 establishes robust consistency priors, the generated outputs may still diverge from complex human perceptual judgments. Conventional refinement approaches often resort to explicit auxiliary losses, but optimizing against these disparate objectives frequently induces gradient conflicts.
To address this, we employ Direct Preference Optimization (DPO). By leveraging holistic preference signals from curated pairs, this phase aims to align the model with human aesthetic and identity standards while strictly maintaining the structural consistency mechanisms acquired in Stage 1.

\subsubsection{Preference Data Construction}

We generate a diverse array of candidate outputs using the Stage 1 model across various prompt-reference scenarios. Prioritizing data quality over quantity, we construct a curated preference pool through rigorous manual evaluation. For each scenario, we identify successful samples (winners $\mathbf{I}_w$) and failed cases (losers $\mathbf{I}_l$) based on a hierarchical evaluation protocol:
\begin{itemize}
    \item Character Consistency: The preservation of fine-grained identity attributes relative to the reference, including facial geometry, hairstyle, and attire.
    \item Anatomical Integrity: The absence of generative artifacts such as malformed limbs or structural incoherence.
    \item Visual Aesthetics: Overall fidelity including lighting coherence, texture detail, and prompt alignment.
\end{itemize}

During training, we employ a combinatorial sampling strategy to dynamically form pairs $(\mathbf{I}_w, \mathbf{I}_l)$ from these annotated pools. This approach effectively expands the training manifold and exposes the model to hard negatives, thereby providing dense and informative gradients for optimization.

\subsubsection{DPO Training Objective}

We instantiate the refinement adapter $\Phi^d$ as a separate, zero-initialized LoRA module. The final policy is parameterized as $\pi_\theta = \Phi + \Phi^c + \Phi^d$, where $\Phi$ represents the frozen pre-trained FLUX weights. Crucially, the Stage 1 consistency adapter $\Phi^c$ remains frozen to preserve the identity-preservation capability, while $\Phi^d$ focuses on aligning the output distribution with human preferences.

Adapting DPO to the flow matching context, we formulate the implicit reward maximization via the difference in matching errors. For a given image pair $(\mathbf{I}_w, \mathbf{I}_l)$, we map them to their corresponding latent representations and sample a timestep $t$ to compute the objective:
\begin{equation}
\mathcal{L}_{\text{stage2}} = -\mathbb{E}_{(\mathbf{c}, \mathbf{Z}_{\text{ref}}, \mathbf{I}_w, \mathbf{I}_l) \sim \mathcal{D}, t} \left[ \log \sigma\left( \beta \left( \log \frac{\pi_\theta(\mathbf{I}_w)}{\pi_{\text{ref}}(\mathbf{I}_w)} - \log \frac{\pi_\theta(\mathbf{I}_l)}{\pi_{\text{ref}}(\mathbf{I}_l)} \right) \right) \right]
\end{equation}
Here, the log-likelihood ratio is approximated by the difference in flow matching errors:
\begin{equation}
\log \frac{\pi_\theta(\mathbf{I})}{\pi_{\text{ref}}(\mathbf{I})} \approx \left\| \mathbf{v}_t - \mathbf{v}_{\text{ref}}(\mathbf{z}_t, t, \mathbf{c}, \mathbf{Z}_{\text{ref}}) \right\|^2 - \left\| \mathbf{v}_t - \mathbf{v}_\theta(\mathbf{z}_t, t, \mathbf{c}, \mathbf{Z}_{\text{ref}}) \right\|^2
\end{equation}
where $\mathcal{D}$ denotes the constructed preference dataset, $\mathbf{z}_t$ is the noisy latent derived from image $\mathbf{I}$, $\mathbf{v}_t$ is the target velocity field, $\mathbf{Z}_{\text{ref}}$ denotes the clean reference latents, and $\beta$ is the temperature parameter.

This formulation encourages $\pi_\theta$ to increase the likelihood of preferred samples by reducing their matching error relative to the reference policy. By explicitly conditioning on $\mathbf{Z}_{\text{ref}}$ in the velocity prediction, we ensure that the preference optimization occurs within the subspace of identity-consistent generation.

\section{Experiments}
\label{sec:experiments}

\subsection{Implementation Details}

\paragraph{Training Configuration}
We conduct all experiments on NVIDIA H800 GPUs utilizing bfloat16 mixed-precision to ensure computational efficiency. For optimization, we employ the AdamW optimizer with specific hyperparameters: $\beta_1=0.9$, $\beta_2=0.999$, $\varepsilon=10^{-8}$, and a weight decay of $10^{-2}$.
We integrate Low-Rank Adaptation (LoRA) into the FLUX.1-dev backbone. Both the rank and alpha ($\alpha$) parameters are set to 16. To ensure comprehensive adaptation, LoRA is applied to all linear layers within the transformer blocks, including query, key, value, and output projections, as well as the feed-forward networks (FFN). 
Input images are resized to an approximate resolution of $1024 \times 1024$ pixels, ensuring dimensions are divisible by the VAE downsampling factor of 16. To mitigate overfitting during training, we apply a caption dropout strategy at a rate of 0.1.

\paragraph{Staged Training Strategy}
The training proceeds in two distinct stages:
\begin{itemize}
    \item \textit{Stage 1:} This stage is trained for 40,000 steps on 4 NVIDIA H800 GPUs with a learning rate of $1 \times 10^{-4}$. The per-GPU batch size is set to 1, resulting in a global effective batch size suitable for establishing identity correspondence.
    \item \textit{Stage 2:} This stage is trained for 4,000 steps on 2 NVIDIA H800 GPUs. We employ a reduced learning rate of $5 \times 10^{-6}$ and set the DPO KL penalty coefficient to $\beta_{\text{DPO}}=1800$. The per-GPU batch size is maintained at 1.
\end{itemize}

\paragraph{Inference Configuration}
During inference, we employ the Euler Discrete Scheduler with 50 denoising steps. To balance identity preservation and text alignment, we set the classifier-free guidance (CFG) scale to 3.5. Crucially, the Group-Shared Attention mechanism remains active during inference to facilitate robust multi-view reference attention.

\paragraph{Benchmark and Metrics}
We evaluate on ViStoryBench~\citep{zhuang2025vistorybench}, which comprises 80 sequences across diverse genres. Following standard protocols, we assess performance using:
\begin{itemize}
    \item \textit{Identity Consistency:} CIDS(Character Identity Score) via ArcFace. We report both \textit{Cross} (vs. Reference) and \textit{Self} (intra-sequence) scores.
    \item \textit{Style Consistency:} CSD (Contrastive Style Descriptors) via CLIP image embeddings to quantify appearance coherence.
    \item \textit{Prompt Adherence:} OCCM (Character Count Matching) and Alignment Score to measure semantic fidelity.
\end{itemize}

\begin{table}[htbp]
  \centering
  \small 
  \setlength{\tabcolsep}{5pt} 
   \caption{Quantitative comparison on ViStoryBench. Our method demonstrates significant improvements in identity and style consistency metrics compared to the best baseline.}
  \begin{tabular}{lcccccc}
    \toprule
    \multirow{2}{*}{\textbf{Method}} & \multicolumn{2}{c}{\textbf{CSD} (Style) $\uparrow$} & \multicolumn{2}{c}{\textbf{CIDS} (Identity) $\uparrow$} & \textbf{OCCM} & \textbf{Alignment} \\
    \cmidrule(lr){2-3} \cmidrule(lr){4-5}
    & Cross & Self & Cross & Self & Score & Score \\
    \midrule
    StoryGen & 37.1 & 53.1 & 37.1 & 51.2 & 80.3 & 41.4 \\
    StoryDiffusion & 34.0 & 54.7 & 39.8 & 58.1 & 84.9 & 67.4 \\
    StoryAdapter & 45.6 & 54.8 & 43.0 & 60.5 & 86.3 & 67.7 \\
    UNO & 39.1 & 60.2 & 45.7 & 60.9 & 89.5 & \textbf{79.6} \\
    \textbf{Ours} & \textbf{57.8} & \textbf{61.8} & \textbf{55.7} & \textbf{63.3} & \textbf{89.6} & 76.9 \\
    \bottomrule
  \end{tabular}
 
  \label{tab:method_comparison}
\end{table}
\subsection{Quantitative Results}
\label{sec:quantitative}

Table~\ref{tab:method_comparison} presents the comprehensive evaluation on ViStoryBench. Our framework demonstrates a commanding performance lead, establishing new state-of-the-art results on 5 out of 6 metrics.

\paragraph{Identity and Style Preservation}
The most significant gains are observed in consistency metrics, validating the efficacy of our progressive two-stage training strategy.
\begin{itemize}
    \item \textit{Character Identity:} In terms of CIDS-cross (fidelity to reference), our method achieves 55.7, surpassing the strongest baseline (UNO, 45.7) by a substantial margin of +10.0. 
    \item \textit{Style Consistency:} We observe a transformative improvement in CSD-cross, where our method scores 57.8 compared to StoryAdapter's 45.6 and UNO's 39.1. This +18.7 leap over UNO indicates that our model successfully decouples identity from realistic priors, enabling robust generation across diverse artistic domains.
\end{itemize}

\paragraph{Semantic Alignment and Robustness}
Regarding prompt adherence, our method achieves an OCCM score of 89.6, effectively matching the strong performance of UNO (89.5), which demonstrates our precise control over character count. 
In terms of Alignment Score, UNO achieves the highest result (79.6), demonstrating its superior capability in semantic interpretation. Our method (76.9) remains highly competitive within a comparable range. The data suggests that our method successfully trades a marginal amount of semantic flexibility ($<3\%$) for a substantial gain in identity preservation (+10.0 CIDS), offering a specialized solution for applications requiring rigid character continuity.

\subsection{Qualitative Analysis and Case Study}
\label{sec:qualitative}

\begin{figure}[t] 
    \centering
    \includegraphics[width=1\linewidth]{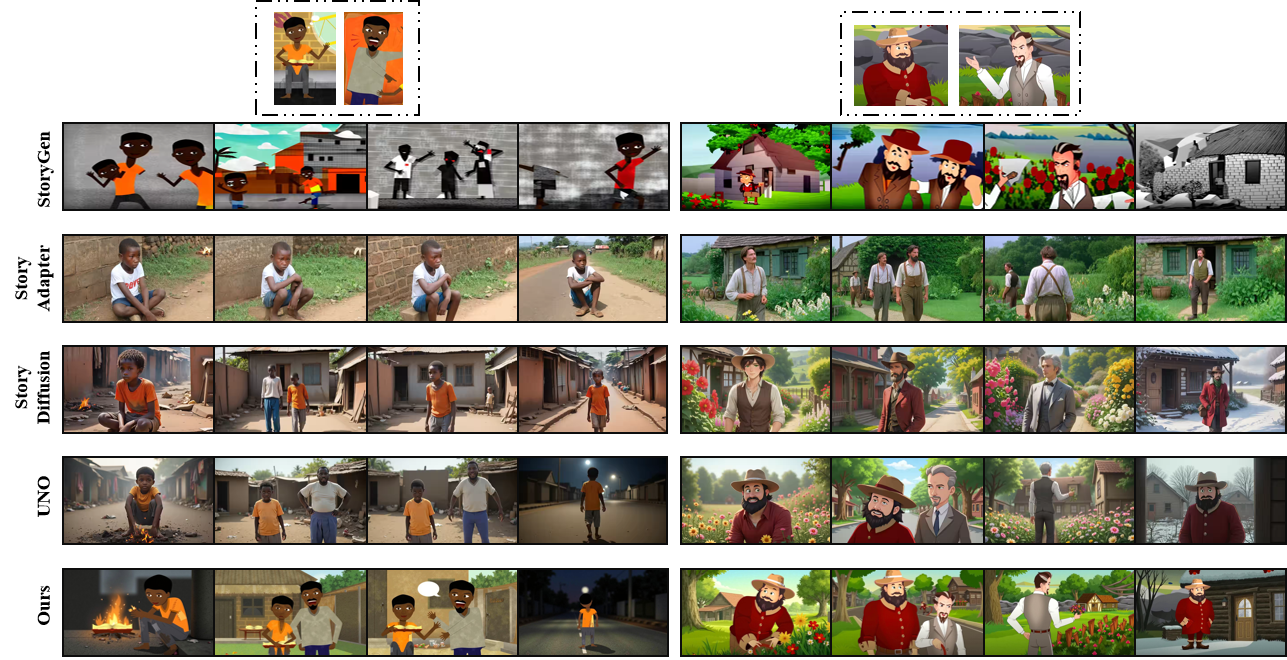}
    \caption{Qualitative comparison on ViStoryBench. Left: A narrative in Flat 2D Illustration style. Right: A narrative in Western Animation style. Unlike baselines that suffer from severe style drift, our method (Bottom Row) faithfully preserves both the artistic style and character identity.}
    \label{figure2}
\end{figure}

We present a detailed qualitative comparison in Figure~\ref{figure2}, illustrating generation results across two distinct narrative genres: a flat 2D illustration style (Left) and a Western animation style (Right). This visual evidence highlights the unique advantages imparted by our Group-Shared Attention (GSA) mechanism, particularly in mitigating style drift and preserving identity consistency in complex scenarios.

\paragraph{Robustness Against Style Drift}
A critical challenge in story visualization is overcoming the realistic bias of pre-trained backbones to adhere to stylized artistic domains. As evident in the Left Panel, the reference inputs dictate a minimalist, flat vector art aesthetic. Existing baselines, particularly StoryAdapter and StoryDiffusion, exhibit severe domain shift. Failing to decouple content from the pre-trained priors, they erroneously render characters as photorealistic humans in cinematic settings. Similarly, StoryGen collapses into severe visual artifacts, losing semantic coherence.Our method (Bottom Row) demonstrates superior style fidelity. It successfully suppresses photorealistic textures to faithfully reproduce the flat shading and color blocks of the reference. This validates that our method effectively retrieves style-specific visual cues from the reference images, ensuring the generated distribution aligns with the target domain rather than the model's internal priors.

\paragraph{Identity Persistence in Dynamic Scenes}
The Right Panel challenges models with multi-character interactions and environmental transitions.
Baseline performance reveals noticeable inconsistency: UNO struggles with facial instability, where the protagonist's facial features and perceived age fluctuate significantly across frames (e.g., between the first frame and the winter scene).
Our model, conversely, maintains both identity and stylistic integrity. It preserves nuanced attributes—such as the specific beard shape and coat style—throughout the sequence, ensuring visual continuity even during complex environmental shifts.

\begin{figure}[t] 
    \centering
    \includegraphics[width=1\linewidth]{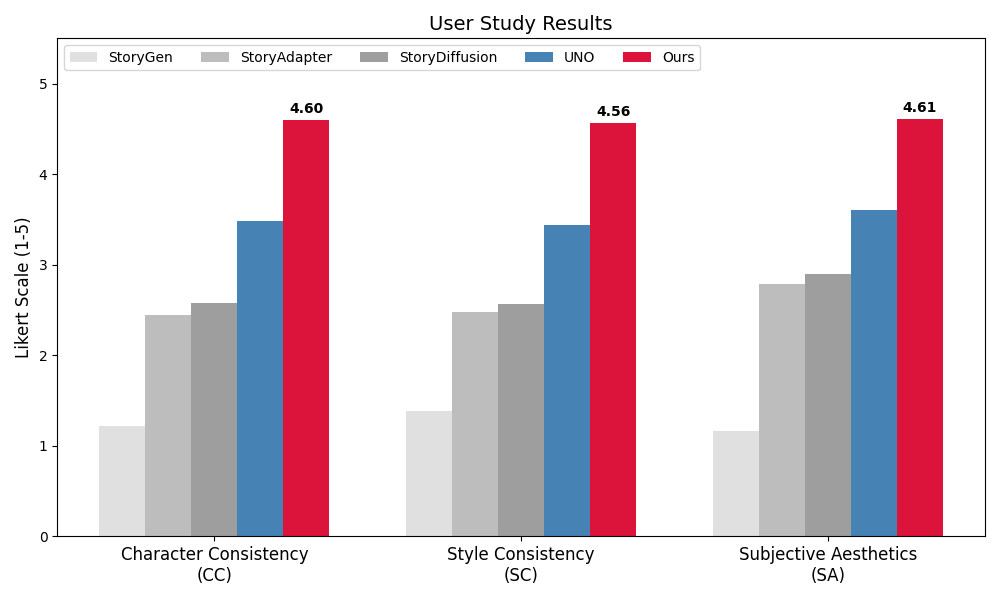} 
    \caption{User Study Results on a 5-point Likert Scale. We compare human preference scores across Character Consistency (CC), Style Consistency (SC), and Subjective Aesthetics (SA). Our method (Red) consistently outperforms all baselines by a significant margin, particularly in identity preservation.}
    \label{fig:user_study}
\end{figure}

\subsection{User Study}
\label{sec:user_study}

\paragraph{Study Protocol}
To validate whether our quantitative superiority translates into perceptible improvements in narrative quality, we conducted a subjective user study with 30 participants. The evaluation set comprised 6 story sequences randomly sampled from ViStoryBench, ensuring coverage across diverse genres. We benchmarked our method against four baselines (StoryGen, StoryAdapter, StoryDiffusion, and UNO) using a randomized, blind evaluation protocol.

\paragraph{Evaluation Metrics}
Participants rated the generated sequences on a 5-point Likert scale (1: Worst, 5: Best) across three perceptual dimensions:
\begin{itemize}
    \item \textit{Character Consistency (CC):} Fidelity of character identity preservation across the sequence.
    \item \textit{Style Consistency (SC):} Coherence of visual style and artistic rendering.
    \item \textit{Subjective Aesthetics (SA):} Overall image quality and narrative appeal.
\end{itemize}

\paragraph{Results and Analysis}
The results are visualized in Figure \ref{fig:user_study}. Our method achieves a commanding lead across all perceptual metrics.
Most notably, in Character Consistency, our method scores 4.60, significantly outperforming the strongest baseline, UNO (3.48). This substantial gap (visually represented by the red bar) confirms that human evaluators perceive our generated characters as much more stable.
Furthermore, in Subjective Aesthetics, our method achieves a score of 4.61, validating that our DPO refinement strategy effectively aligns the model with human aesthetic preferences, producing stories that are not only consistent but also visually captivating.

\begin{table}[htbp]
  \centering
  \small
   \caption{Ablation study of key components on ViStoryBench. We quantitatively evaluate the specific contributions of Group-Shared Attention (GSA) and DPO refinement to identity (CIDS) and style (CSD) preservation.}
  \begin{tabular}{lcccc}
    \toprule
    \multirow{2}{*}{\textbf{Method}} & \multicolumn{2}{c}{\textbf{CIDS} (Identity) $\uparrow$} & \multicolumn{2}{c}{\textbf{CSD} (Style) $\uparrow$}\\
    \cmidrule(lr){2-3} \cmidrule(lr){4-5}
    & Cross & Self & Cross & Self \\
    \midrule
    w/o GSA & 34.4 & 53.6 & 29.6 & 60.6 \\
    w/o DPO & 46.4 & 58.2 & 46.7 & 56.6\\
    \textbf{Full Method} & \textbf{55.7} & \textbf{63.3} & \textbf{57.8} & \textbf{61.8}  \\
    \bottomrule
  \end{tabular}
    \label{tab:ablation}
\end{table}

\subsection{Ablation Study}
\begin{figure}[htbp] 
    \centering
    \includegraphics[width=1\linewidth]{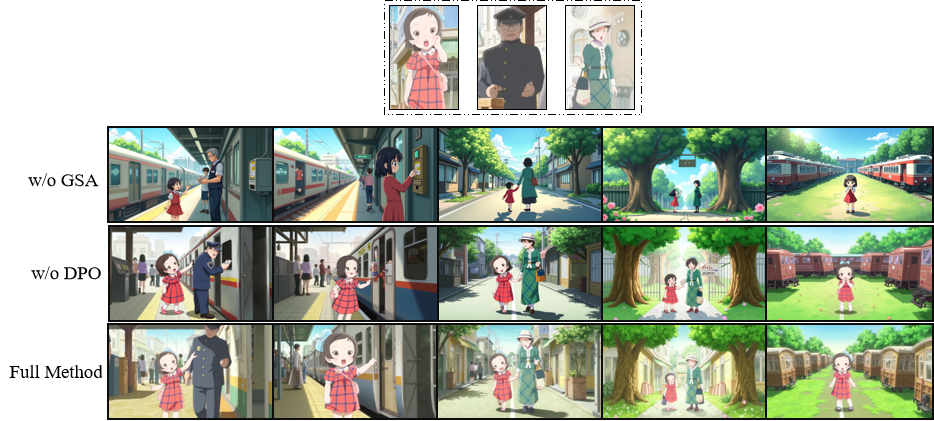}
    \caption{Visual ablation study. 
    Top (w/o GSA): Removing GSA causes severe drift in both style and identity. 
    Middle (w/o DPO): Removing DPO results in lower fidelity and anatomical artifacts. 
    Bottom (Full Method):Our approach achieves the best consistency and detail sharpness.}
    \label{fig:ablation}
\end{figure}

To validate the efficacy of our framework, we conduct a component-wise analysis to disentangle the contributions of our two core modules: the Group-Shared Attention (GSA) and the DPO refinement strategy. We evaluate the impact of removing each component while keeping other settings constant. The quantitative results are detailed in Table \ref{tab:ablation}, and visual comparisons are presented in Figure \ref{fig:ablation}.

\paragraph{Necessity of Group-Shared Attention (GSA)}
We first investigate the structural foundation of our model by replacing GSA with standard self-attention (denoted as w/o GSA). As shown in Table \ref{tab:ablation}, this removal leads to a catastrophic drop in identity preservation, with CIDS-cross from 55.7 to 34.4. This sharp decline confirms that standard diffusion attention, which is restricted to local tokens within a single image, is fundamentally incapable of maintaining identity across different generation sessions. Without GSA to retrieve features from the reference cache, the model treats each frame independently, leading to severe identity hallucination and inconsistent clothing (as seen in Figure \ref{fig:ablation}, Top Row). Thus, GSA serves as the indispensable backbone for structural consistency.

\paragraph{Effectiveness of DPO Refinement}
We further assess the value of the second training stage by evaluating the model trained only with Stage 1 (denoted as w/o DPO). While the Stage 1 model establishes basic consistency, adding DPO refinement yields a substantial performance boost, increasing CIDS-cross by +9.3 and CSD-cross by +11.1. Qualitative results (Figure \ref{fig:ablation}, Bottom Row) reveal that while Stage 1 captures the rough identity, it often suffers from semantic drift or anatomical artifacts . The DPO stage rectifies this by aligning the model with expert-curated preference pairs. This process acts as a semantic filter, forcing the model to reject low-fidelity generations and sharpening fine-grained details, thereby bridging the gap between rough structural alignment and high-fidelity perceptual quality.

\section{Conclusion}

In this paper, we proposed a novel two-stage framework for character-consistent story visualization, integrating architectural specialization with preference-based alignment. Our approach addresses the core challenge of maintaining robust character identity and narrative coherence across dynamic scenes while ensuring strict fidelity to textual descriptions.

Our empirical findings validate two key design principles. First, we demonstrate that structural consistency mechanisms offer a superior alternative to traditional loss-based constraints. By introducing Group-Shared Attention (GSA), we structurally encode consistency into the generation process, enabling the model to intrinsically learn cross-frame correspondences via direct access to co-batch visual tokens. Second, we establish that Direct Preference Optimization (DPO) serves as a critical refinement stage. Functioning as a holistic optimization objective, DPO aligns the model with human aesthetic judgments, effectively resolving the trade-offs between visual fidelity and identity preservation. This synergy yields generated narratives that are characterized by both high perceptual quality and strict logical coherence.

Through this modular strategy, we demonstrate that architectural priors and preference optimization are highly complementary. This decoupled pipeline offers a scalable and effective paradigm for future research in consistent multi-sample generation.

\noindent\textbf{Limitations and Future Work.} Despite its effectiveness, our method has certain limitations. The computational cost of GSA scales with the number of reference images, which may constrain real-time applications for extended narrative sequences. Furthermore, maintaining consistency for highly abstract or non-anthropomorphic characters remains a non-trivial challenge. Future work will focus on optimizing the attention mechanism for linear complexity and extending our framework to consistent video generation tasks. We believe this work paves the way for more controllable and coherent generative storytelling.

\clearpage






\end{document}